\documentclass{article}
\usepackage{arxiv}
\usepackage{amsmath,amsfonts}

\usepackage[utf8]{inputenc} 
\usepackage[T1]{fontenc}    
\usepackage{hyperref}       
\usepackage{url}            
\usepackage{booktabs}       
\usepackage{amsfonts}       
\usepackage{nicefrac}       
\usepackage{microtype}      
\usepackage{lipsum}
\usepackage{graphicx}
\graphicspath{ {./images/} }

\usepackage{subcaption}
\usepackage{algorithm}
\usepackage{array}
\usepackage{textcomp}
\usepackage{stfloats}
\usepackage{url}
\usepackage{verbatim}
\usepackage{cite}
\usepackage{algpseudocode}
\usepackage{xcolor}
\usepackage[normalem]{ulem}
\usepackage{changepage}

\begin{document}
\title{Guidelines For The Choice Of The Baseline in XAI Attribution Methods\thanks{
    This work has been submitted to the IEEE for possible publication. Copyright may be transferred without notice, after which this version may no longer be accessible.
    
    Cristian Morasso, Ilaria Boscolo Galazzo, and Gloria Menegaz are with the Department of Engineering for Innovation Medicine, University of Verona, Verona, Italy.
    
    Giorgio Dolci is with the Department of Computer Science, University of Verona, Verona, Italy, and the Department of Engineering for Innovation Medicine, University of Verona, Verona, Italy.
    
    Sergey M. Plis is with Tri-Institutional Center for Translational Research in Neuroimaging and Data Science, Georgia State University, Georgia Institute of Technology, Emory University.
    
    Corresponding author: Cristian Morasso (cristian.morasso@univr.it)
    
    This work was partially supported by the Ministero dell'Università e della Ricerca (Bando Progetti di Rilevante Interesse Nazionale, PRIN 2022, "AI4BRAVE: AI for modeling of the Brain-Heart Axis in aging", project-reference code 202292PHR2) as well as the NIH grant R01MH129047.}}

\author{Cristian Morasso, Giorgio Dolci, Ilaria Boscolo Galazzo, Sergey M. Plis, Gloria Menegaz}


\maketitle

\begin{abstract}   

Given the broad adoption of artificial intelligence, it is essential to provide evidence that AI models are reliable, trustable, and fair.
To this end, the emerging field of eXplainable AI develops techniques to probe such requirements, counterbalancing the hype pushing the pervasiveness of this technology.
Among the many facets of this issue, this paper focuses on baseline attribution methods, aiming at deriving a feature attribution map at the network input relying on a "neutral" stimulus usually called "baseline".
The choice of the baseline is crucial as it determines the explanation of the network behavior. In this framework, this paper has the twofold goal of shedding light on the implications of the choice of the baseline and providing a simple yet effective method for identifying the best baseline for the task. To achieve this, we propose a decision boundary sampling method, since the baseline, by definition, lies on the decision boundary, which naturally becomes the search domain. Experiments are performed on synthetic examples and validated relying on state-of-the-art methods. Despite being limited to the experimental scope, this contribution is relevant as it offers clear guidelines and a simple proxy for baseline selection, reducing ambiguity and enhancing deep models' reliability and trust.

\end{abstract}


\section{Introduction}
\label{sec:introduction}

Saying that Artificial Intelligence (AI) pervades all the fields of life and sciences is no more a news. However, behind the hype pushing its exploitation across the different domains, are inherent risks that need to be carefully identified and faced. This is exacerbated by the advent of deep learning (DL), for the inherent complexity of the models that goes far beyond the limits of human comprehension. Not understanding pushes trust in opposite directions: blind acceptance by one side, reflecting the excess of trust, and refusal of the adoption of AI technologies by the other, expressing lack of trust. This last permeates the biomedical field, where erroneous AI-driven actions could lead to dramatic consequences.
To mitigate this, eXplainable AI (XAI) methods are being increasingly adopted in order to decode the models' reasoning and interpret the outcomes, leading a step forward along the path to trustworthiness. However, many challenges remain and need to be faced to provide evidence of the soundness, reliability, and trustworthiness of the XAI outcomes. This concern is particularly critical in the biomedical field, where ground truth is often lacking, and the underlying involved processes are complex, interdependent, and often still unknown, making the validation of XAI outcomes particularly complex. 
Just to give an example, there is no guarantee that different XAI methods applied to a given task would provide consistent results \cite{ancona2018towards, kindermans2019reliability}. On the contrary, this is most often not the case because the outcomes depend on the models, on the XAI method itself, and on the data, such that the overall level of uncertainty can be, in general, decreased yet not annihilated. In consequence, the requirements of soundness, reliability, and trustworthiness of the XAI go hand in hand with the need of validation methods ensuring the respect of such attributes.

Among the different methods, gradient-based methods are most widely spread for the inherent interpretability of Neural Networks (NN) relevance or attribution maps. Visual stimuli are very natural to interpret and do not require the mediation of language constructs for conveying the message, making them an effective communication means. However, trust must rest on solid assumptions ensuring the reliability and robustness of the outcomes. 

Within the many options that are available at the state-of-the-art (SOA) for generating relevance maps, here we focus on gradient-based methods and more specifically on those methods that produce contrastive maps with respect to a "neutral" stimulus, usually called "baseline" (BL). Behind the apparent simplicity of such methods hides the complexity that must be acknowledged and solved to ensure the meaningfulness of the relevance map. The choice of the BL is particularly critical in this respect and is the core of this contribution. Being a neutral stimulus for the network, the BL sits on the decision boundary (DB), and hence, the network has no preferences in classifying the input. In consequence, the identification of the DB is essential for the choice of a BL.

In what follows, the SOA is presented, touching different relevant topics, which are i) XAI methods, ii) Baseline Attribution Methods (BAMs), and iii) DB identification in NN.
\subsection{Explainable AI}
Explainable AI is a branch of AI aiming at decrypting and revealing what the so-called \textit{black-box} learned \cite{SAEED2023110273, weber2023beyond}, to allow humans to understand what AI algorithms apprehended, projecting the knowledge learned by the AI system in a human-compatible space, and explaining its behavior and characteristics. 
During the last few years, a variety of XAI algorithms have been proposed, relying on different hypotheses and providing different types of explanations \cite{rahman2023looking, minh2022explainable, tjoa2020survey, munroe2024applications}. 
Focusing on NN, the most widespread post-hoc methods, that is LIME \cite{ribeiro2016should} and SHAP \cite{lundberg2017unified}, are not the best choice because of many reasons including the large feature and sample space, the inherent limitations in dealing with correlated data, and the difficulty of coping with models allowing diverse and often missing data, requiring multi-view, multi-task and/or generative models.
As already mentioned, feature visualization methods providing an attribution map at the network input are often preferred due to the ease of interpretation and application. Indeed, many tools have been made available and gained popularity, pushing the adoption of these methods that are often used as a silver standard in the SOA.

\subsection{Baseline attribution methods}
Baseline Attribution Methods (BAM) \cite{lundstrom2022rigorous} rely on a special hyperparameter, the BL, used for computing the relevance of the input features.
The BL is a neutral input for the NN and as such sits on the DB. Being oriented to computer vision and multimedia applications, in the literature the notion of {\em neutrality} is most often cast on perceptual cues. In consequence, all-zero images or instances of different kinds of noise (e.g., Gaussian or white noise) that are meaningless for the human observer, were used as BLs \cite{sturmfels2020visualizing, haug2021baselines}. Sometimes, blurred images were also considered \cite{sturmfels2020visualizing, sundararajan2017axiomatic}. If this provides a simple, easy to implement and intuitive solution, care must be taken when interpreting the resulting attribution map, even though the sanity check for neutrality is satisfied \cite{sundararajan2017axiomatic}.
Indeed, even though the signal is neutral for the network, different BLs can lead to very different attributions \cite{mamalakis2023carefully,dolci2023objective, arras2022clevr, haug2021baselines, izzo2020baseline}, limiting their adoption in critical domains. If this led to the awareness of the feature map and BL dependency, on the other hand, no general solutions have been proposed so far, grounded on solid assumptions, leading to stable solutions that ensure consistency and contrastiveness across different input samples. 

In literature, a variety of BLs for BAM methods have been proposed for computing features' attributions \cite{shrikumar2017learning, chen2021explaining, sundararajan2017axiomatic} also facing the attributions' dependency on the BL. 
In \cite{dolci2023objective} Dolci et al. employed Integrated Gradients (IG) \cite{sundararajan2017axiomatic}, a BAM method, for providing evidence of the feature attribution dependence from the BL in the neuroimaging context. To this end, several BLs were considered, including black matrices, Gaussian noise, uniform noise, and mixed noise, in a classification task. Mamalakis and colleagues \cite{mamalakis2023carefully} used a geoscientific domain comparing data-derived BLs through IG and DeepSHAP \cite{chen2021explaining}, obtaining different attributions with different BLs. Shi et al. \cite{shi2022output} used the Aumann–Shapley XAI method \cite{aumann2015values} to compare signals such as Gaussian noise, black matrix, and max distance, obtaining different attributions in a classification task, using ImageNet. Izzo and colleagues \cite{izzo2020baseline} used a credit card-related dataset and synthetic data for comparing the full zero input (black matrix for tabular data), max distance, and a boundary BL, which is a BL that lies on the DB of the classifier. 
Kindermans et al. \cite{kindermans2019reliability} compared different XAI methods, including IG, and showed the impact of different BL usage on the attribution map, highlighting the unreliability of the zero BL, and concluding that the choice of the BL must account for the task and the application domain. However, a clear guidance to the solution was not proposed by the authors.
Arras et al. \cite{arras2022clevr} proposed a framework performing a similar comparison, and found that using a domain-related BL (mean image and channel mean values) led to improved results, as opposed to using the zero BL that led to poor and unreliable results.
Sturmfels et al. \cite{sturmfels2020visualizing} proposed a comprehensive analysis of several BLs, such as zero, Gaussian and uniform noise, blurred images, and max distance. They concluded that none of these BLs was the optimal choice for the given task.
Ancona et al. \cite{ancona2018towards} stated that the BL aims to simulate the absence of a feature and to do that, the BL must necessarily be chosen in the domain of the input space, and this inherently creates ambiguity between an ordinary input that has equal features of the BL and a placeholder for that features.

In order to obtain more accurate attributions, the common practice consists of retaining only the absolute value of the attributions \cite{hooker2019benchmark, sturmfels2020visualizing}, hence relying only on the feature impact \cite{sturmfels2020visualizing}.
However, the possibility of relating the sign of an attribution value to the indication of the increase/decrease of the corresponding feature with respect to the BL would bring a step forward in the interpretation process. In the biomedical field, this would provide indications for the identification of new biomarkers and empower biomarkers' discovery.
Ensuring that the sign of the attribution conveys information about how a change (increase/decrease) in the value of the feature pushes the sample towards the target class is essential for the interpretation of the outcome. To give an example, if the feature represents the volume of the hippocampus in a classification task aiming at segregating healthy controls from Alzheimer's patients, an increase in the feature bringing to the healthy class should be reflected in a positive attribution value, while a negative one would be an indication of disease. In consequence, the possibility to assign a clear and unambiguous meaning to the sign of the attributions becomes essential.

\subsection{Decision boundary identification in neural networks}
As mentioned above, being a neutral signal to the network, the BL must sit on the DB, reflecting the fact that it must result in a similar score across the classes \cite{lee1997decision}. Therefore, accurately determining the DB is essential for generating neutral BL samples. Consequently, the challenge shifts to identifying the DB, which will then serve as the basis for either sampling or validating the BL signal.

In their pioneering work tracing back to 1997, Lee et al. \cite{lee1997decision} seeded this perspective, showing that from the DB, it is possible to extract all feature attributions relying on the boundary gradient, providing evidence that locating the DB is crucial for comprehending the role of the features in the task under analysis.

The problem of the identification and characterization of the DB has been deeply investigated in the literature, though the advent of deep learning added levels of complexity. Karimi et al. \cite{karimi2019characterizing} exploited DB to define a metric for computing the complexity of their task, based on the non-linearity of the DB. Oyallon \cite{oyallon2017building} analyzed the hyperparameters' impact on the DB, while Lee et al. \cite{lee1993feature} proposed a technique for using the DB as a feature extractor to derive the features' contributions. 

So far, DB has been mainly used for extracting the region of influence of each class, also called Decision Boundary Maps (DBMs) \cite{rodrigues2018image,rodrigues2019constructing}, most often following the same pipeline. First, a dimensionality reduction is performed through an operator bringing to the two-dimensional space, enabling the visualization. 
Then, the projection domain is sampled at regular intervals \cite{benato2024human}. Finally, the samples are projected back to the original space and classified to obtain their labels. 

Another way of approaching the DB identification relies on adversarial attacks, exploiting feature sampling or perturbations in order to fool the NN or boost its robustness.
The proposed algorithms are mainly based on the gradient of the input \cite{goodfellow2014explaining, heo2019knowledge,jiang2020searching,tian2023knowledge} using both gradients and binary search to identify the DB. However, these suffer from low accuracy and precision in the DB localization, also known as Data Manifold closeness \cite{verma2020counterfactual}.
The exploitation of adversarial perturbation to explain NN is known as counterfactual explanations \cite{wachter2017counterfactual,freiesleben2022intriguing}.
Following this idea, Moore et al. \cite{moore2019explaining} proposed a solution that involves the addition of constraints to force the gradient in the desired direction. Their task consisted in explaining, via counterfactual explanation, why a loan had been rejected by an NN and what a customer should do to change the model output. For instance, they proposed the customer's age as a constraint, so the counterfactual explanation must include the other features, such as the salary. However, this method requires human intervention, limiting its generalizability.

Only a few algorithms compute the DB or sample it without involving the gradient. In \cite{guan2020analysis} the authors sample at regular intervals along the straight line that connects two instances of different classes in order to detect those lying on the DB. He et al. \cite{he2018decision} perturbed the input along random directions by a fixed step until the perturbed input was assigned to a different class. Karimi and colleagues \cite{karimi2019characterizing} employed a different approach in which two autoencoders were used to generate two samples close to the DB (with different classification outcomes), and then looked for an additional sample on the line that linked the two generated samples, similar to \cite{guan2020analysis}.

Recent methods gaining popularity related to DB detection and visualization are SplineCAM \cite{humayun2023splinecam} and DeepView \cite{schulz2019deepview}.
These are particularly interesting in that they allow the DB of an NN to be traced at its input in a two-dimensional projection. 
SplineCAM \cite{humayun2023splinecam} is a SOA method used to analyze the NN's geometry. SplineCAM is the first exact method that is fast and scalable and computes the partition on a two-dimensional space that corresponds to a projection of a DB of NN.
DeepView \cite{schulz2019deepview} relies on a dimensionality reduction method that allows visualization, exploiting the UMAP method \cite{mcinnes2018umap} and, thanks to a generalization of the Fisher metric, it is able to perform the projection to a lower dimensional space (e.g., two-dimensional), providing a human-interpretable mapping of the DB.

\subsection{Paper contributions}
In this work, we aim at providing guidance for the choice of the BL relating it to the identification and exploitation of the DB in NN. To this end, we focus on the IG method and we propose the Informed Baseline Search (IBS) algorithm, allowing i) to identify the DB, providing the search domain for BL candidates, where the optimal BL for any input samples can be picked, and ii) to select the optimal BL for each input sample following a predefined criterion. In detail, IBS first samples the DB following a simple yet effective iterative search procedure, and then selects the best DB sample as the BL for the given input sample. 
The reasons behind the choice of IG rely on the ease of application and interpretation of the outcomes. It is a well-known method that has been extensively used relying on different BLs for computing attributions, including zeros \cite{goh2021understanding, sundararajan2017axiomatic,sturmfels2020visualizing, mamalakis2023carefully,sayres2019using, ancona2018towards}, different forms of noise \cite{goh2021understanding,sturmfels2020visualizing}, and blurred images \cite{sturmfels2020visualizing}. 

The proposed method for DB detection was validated by assessing the consistency of the results with two SOA methods, SplineCam and DeepView.

Then, given the DB, a criterion for picking the optimal BL was devised and illustrated relying on low-dimensional toy examples showing the idea and the rationale in an intuitive way. Finally, the input-sample-driven BL selection advantages are discussed by a post-hoc analysis contrasting the XAI attributions extracted by optimal and sub-optimal BLs in light of the implications of the subsequent interpretation.

In summary, the main contributions of this work are:
\begin{itemize}
\item Analysis of the implications of the choice of the BL in BL-guided XAI methods;
\item Casting the BL identification problem in a DB detection framework;
\item Providing a simple proxy for the DB detection at the network's input, called Informed Baseline Search (IBS);
\item Definition of an optimality criterion for the choice of the BL enabling unambiguous interpretation and ensuring contrastiveness;
\item Experimental validation of the proposed algorithm on synthetic data.
\end{itemize}
The Informed Baseline Search is made publicly available at \url{ https://github.com/BraiNavLab/informed_baseline_search} as well as the code needed to generate the synthetic data.

\section{Background}
\subsection{Integrated Gradients}
IG \cite{sundararajan2017axiomatic} is one of the most popular BAM algorithms. The basic idea consists in accumulating the gradients along the path linking the BL with the sample under analysis and weighting the results by the distance between the two.

Formally, consider $f: R^{n} \longrightarrow [0,1]$ and $x \in R^n$ as the input and $x' \in R^n$ as the BL, IG computes the input attribution by accumulating gradients along the $x - x'$ line; due to that, IG is defined as a path method and formulated as follows:
\begin{equation}
    \label{formula:IG}
    IG_{i}(x) := \left(x_i - x_i' \right) \times \int_{\alpha=0}^1 \frac{\partial f\left(x' + \alpha \times \left( x_i - x_i' \right) \right)}{\partial x_i} d\alpha 
\end{equation}

IG is based on two fundamental axioms that are sensitivity and implementation invariance. The first asserts that if the change of one feature, of the input or the BL, leads to a different prediction, that feature attribution would be given a non-zero attribution. 
In addition, if the network does not depend on some variables, their attributions should always be zero.
The latter means that the attributions are always identical for two functionally equivalent networks.

Following \cite{ancona2018towards}, 
IG can perform both global and local feature (input) attribution, describing, respectively, the influence of a feature on the output with respect to a BL and the output modification induced by infinitesimal feature perturbations.

\label{Bg:IG_delta_bs}
As denoted in Equation (\ref{formula:IG}), IG can be decomposed into two elements, i) the \emph{Delta} that is the distance between input and BL ii) the \emph{Cumulated Gradients (CG)}, that is the accumulation of the gradients of all points along the IG path. While the CG provides the mean of the gradient distribution along the path, expressing the overall impact of the features for the target class, the Delta weights such a value sample-wise, giving more weight to the farthest samples.

\subsection{SplineCAM}
SplineCAM \cite{humayun2023splinecam} is used to analyze the network geometry by exploiting splines. 
A spline is a mapping to a degree $D$ polynomial of on each region $\omega$ of its input space partition $\Omega^l$ of the layer $l$, with the additional constraints that the first $D -1$ derivatives of those polynomials are continuous throughout the domain.
Considering the geometry of a trained network, the extraction of the DB can be relatively easy.
To do that, SplineCAM authors first proposed to compute the single-layer $l$ DBs (one hyperplane for each neuron in the layer) by exploiting the $l-1$ output as input and analyzing the $l$ activation. Secondly, they defined a projection to back-project the $l$ DBs in the $l-1$ input, hence, to input space, iterating from $l-1$ to $0$. Then, they run two algorithms (\textit{find partitions} and \textit{find cycles}). The first one, given the DBs of layer $l$ back-projected to the input, computes finer partitions and creates a graph $G$ with them. The latter extracts unique cycles in $G$, obtaining $\omega \in \Omega^l$
SplineCAM repeats this process for all layers up to $l=L$ to obtain the final partition and so the boundary. For more details we refer to the paper \cite{humayun2023splinecam}.

Regarding the limitations, SplineCAM is able to compute DB only in two dimensions; it is possible to reduce the problem with a projection if the feature number is limited. However, reducing a task into two dimensions is not trivial.
Another notable limitation of SplineCAM is the limited set of NN layers available, i.e., it can not deal with max-pooling, a core operation for convolutional NN (CNN).

\subsection{DeepView}
DeepView \cite{schulz2019deepview} is a DBM method that exploits UMAP \cite{mcinnes2018umap} for the sake of visualization. Thanks to a generalization of the Fisher's metric (DiDi technique), it is able to perform the transformation $\pi$ from an input $x$ to a lower sample $y$ with two dimensions.
UMAP computes the $\pi$ projection with a custom distance; it is composed of two different components, the first one to track the spatial distance $d_S$ (Euclidean or Riemannian) and the latter to track the distance in the prediction of the network $f$ called $d_JS$, or \emph{pullback-metric}.
The \emph{pullback-metric} is an indicator for the boundary features in the sense that it tells whether two points are mapped to the same class or not. Additionally, the authors proposed the inverse projection $\pi^{-1}$ able to map two-dimensional data into original space. Finally, the pipeline of DeepView is the following: i) apply the Fisher UMAP technique to find a projection $\pi$ to map original data into two-dimensional data $y_i = \pi(x_i)$; ii) create a tight regular grid of samples $r_i$ in the two-dimensional space and map it to the high-dimensional space using the inverse projection $\pi^{-1}$, $s_i = \pi^{-1}(r_i)$; iii) classify $s_i$ with $f$ to obtain predictions; and iv) visualize the two-dimensional space with the classification.

\section{Problem statement}

\label{sec:prob_def}

In this Section, the main criticalities of choosing the BL are analyzed, starting from the impact of the BL location and distance to the sample in simple toy examples.

The first critical point of the trivial baselines is related to their distance and location with respect to the data distribution, as already discussed in the literature \cite{kindermans2019reliability, arras2022clevr, izzo2020baseline, ancona2018towards}. Considering IG, the gradients are "summed" from the BL to a specific input, so that when the BL falls far from the data distribution, the value of the attribution would not be reliable.
In this case, the attributions would be just a scaled version of the gradients' integration, where the scaling factor is the distance of the sample to the BL, and very similar for samples belonging to the two classes, respectively. 

The most common practice for dealing with this problem is to keep only absolute values; this procedure can be appropriate in general computer vision tasks, where the aim is to extract the main features and not understand the effect of their modulation in the classification outcomes of a given sample. On the contrary, in critical domains, such as biomedical and clinical ones, such a trivial solution could not be suitable because understanding the effect of increasing or decreasing a feature is essential for deriving biomarkers and/or making diagnostic decisions.

Our proposal is to sample the {\em inner} DB to extract meaningful BLs, where {\em inner} DB is defined as the portion of DB that lies in the data manifold, hence resulting in the informative portion of DB able to explain the network \cite{lee1997decision}.

\begin{figure}[!t]
    \centering
    \includegraphics[width=.5\linewidth]{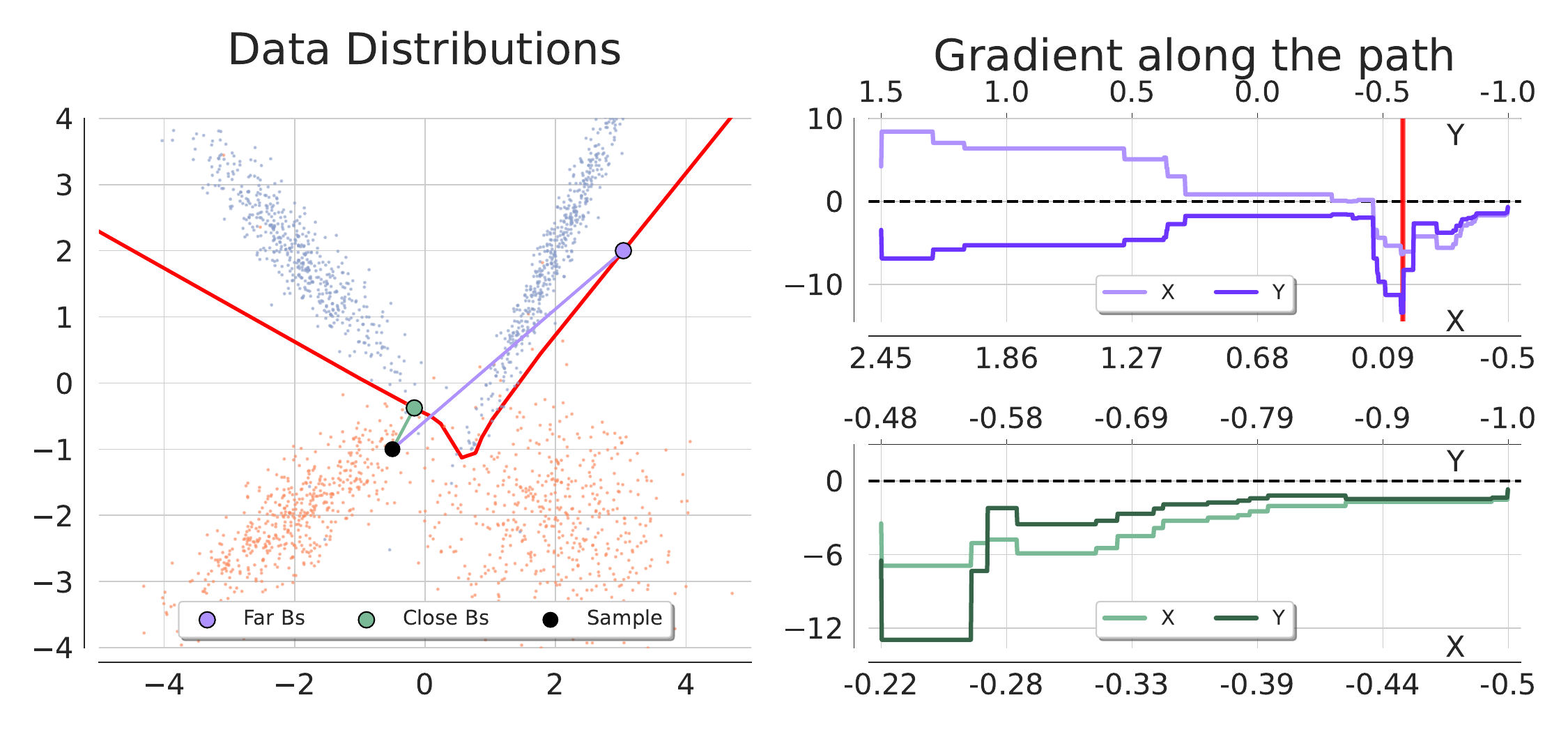}
    \caption{Gradients behavior along the interpolated path between two different BLs, the purple one is randomly sampled on the DB, while the green BL is orthogonal to the sample. The red vertical line in the purple plot represents the location of the crossed DB. The lower axis of the line plots represents the X value, and the top axis represents the Y value.}
    \label{fig:grad_comp}
\end{figure}

In what follows, we refer to the result of the path-wise interpolation of the gradients as Cumulated Gradients (CG), and to the length of the path from the BL to the sample as Delta ($\Delta$). Fig. \ref{fig:grad_comp} illustrates a simple toy example.
In the plot at the left, let the green and purple dots represent two BLs lying on the DB (red line), and the black dot represents the sample of interest, while those on the right show the value of the gradients with respect to the two features, sampled along the path from each of the two BLs to the sample, respectively.
Analyzing the gradient signal of the closer BL (green dot), the gradient absolute value peaks close to the BL, and then decreases, as expected, as it approaches the sample. 
The gradient magnitude along the path from the purple dot to the sample presents two peaks: the first at the baseline location and the second at the intersection between the path and the boundary.
Of note, the pattern of the gradient signal at the crossing point is similar to the one of the green BL, which was to be expected since the same DB segment is being crossed.
Due to the crossing, the result of the gradient integration across the purple path is not reliable and would lead to meaningless attributions. This would apply to any BL leading to a path crossing the DB, highlighting the fact that the baseline-to-sample path must not cross the DB for deriving useful CGs. 

To avoid this from happening, the proposed solution consists in choosing the {\em closest} boundary point to the investigated sample as the BL. 
Since the gradient is orthogonal to the boundary \cite{lee1997decision}, and following the definition of the $\Delta$ factor, this will lead to positive attribution values. In consequence, the CG and $\Delta$ contributions to the IG must be analyzed separately to preserve the information conveyed by the sign.

To summarize, we define the \textit{optimal} BL as a sample that holds the following properties: i) being neutral to the network, thus lying on the DB; ii) sitting within the data distribution domain; and iii) being the closest sample of the DB to the datapoint under analysis. For this last property to hold, the BL sample should be the orthogonal projection of the datapoint on the DB. However, if this condition can not be met, the sub-optimal BL is the closest point of the DB to the sample should be considered ($\min_{x' \in DB} (\| (x - x')\|^2)$).

The first two conditions are essential to ensure that the BL lies in the portion of the feature space that was used to train the network, while the latter ensures positive attributions and prevents crossing multiple DB segments with the optimal BL, the sub-optimal approach only preserves the constraint of not crossing multiple DB segments.

The proposed approach brings interesting properties, with the inherent difficulty of computing or at least sampling the DB not being trivial. In what follows, we present a simple yet effective boundary sampling method leading the DB detection in the region of interest, that is, within the data manifold. After that, the user can select the optimal one for its investigated sample, according to the above-mentioned guidelines.

\section{Informed Baseline Search} 
In order to solve this problem, a constrained problem has been formulated. Let $F(\textbf{x}$) be the model, $C=2$ the number of classes, $\Omega$ the data distribution domain, $X_{\Omega} \in \Omega$ the DB sampling domain and $\textbf{x}$ be the result of the sampling. The search rule is then
\begin{align*}
    & \text{Find:} & \textbf{x} : P(\mathcal{F}(\textbf{x})) = \frac{1}{2}\\
    & \text{Subject to:}& \textbf{x} \in X_{\Omega}
\end{align*}

\subsection{Algorithm}
\begin{figure}[!t]
    \centering
    \includegraphics[width=.5\linewidth]{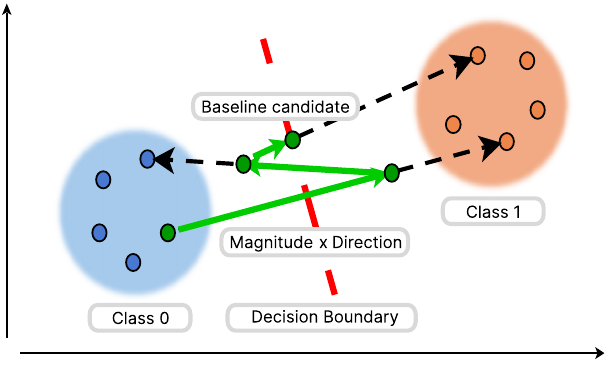}
    \caption{Two feature graphic example of the algorithm (three steps), the green star in the blue class (0) is the BL search starting point, and the red dotted line represents the DB. Running IBS on the starting point, as first, the losing class is 1, so it extracts the \textit{direction} from \textit{BL} to \textit{Class 1}, then moves the \textit{BL} over this direction for a given \textit{magnitude} (green vector), in order to obtain the next BL candidate, then the prediction is in favor of class 1, so it does the same with class 0, so on and so forth until an optimal BL is reached. }
    \label{fig:alg_graph}
\end{figure}

The basic idea is to iteratively sample from the class distributions and to check the class probabilities of the computed sample. The process continues until equal probability across classes is reached. Figure \ref{fig:alg_graph} illustrates the case $C=2$ (two classes task).

Given the target distributions $T_0, T_1$ that represent the two classes and a starting point, the algorithm computes the class predictions and then starts moving with prediction-related magnitude to the opposite class, pointing to a randomly sampled point. The algorithm iterates these steps until the DB is reached; therefore, a probability of $0.5$ is reached for both classes within a predefined precision.
The algorithm is based on sampling from target distributions to remove constraints that limit the ability to sample the majority of the inner DB, such as restricting the algorithm to search the DB only along the line connecting two known samples. However, fixing $T_0 = \{x_0\}$, $T_1 = \{x_1\}$ and starting the search from a point on $[x_0,x_1]$ line will force the algorithm to find the DB sample on the line.

\begin{algorithm}
\caption{Informed Baseline Search}\label{alg:BSA_code}
\begin{algorithmic}
\Require $F^a$ as NN function with activation function $a$, $bl$ the BL, $x$ a sample of a class, $T_c$ the target of class $c$ and $sf$ the scaling function.
\State $bl \gets x$
\State $predict \gets F^a(bl)$ 
\State $step \gets 0$
\While{$predict \neq 0.5$}
\State $predicted_{class} \gets int(predict) $
\State $target \gets T_{1}$ if $predicted_{class}$ else $T_{0}$
\State $direction \gets target-bl$ 
\State $magnitude \gets \|predict - 0.5\|*sf(step)$
\State $bl \gets bl+ direction * magnitude$
\State $predict \gets F^a(bl)$ 
\State $step \gets step +1$
\EndWhile
\State \Return $bl$
\end{algorithmic}
\end{algorithm}

The algorithm has been developed such that if the starting point is in the feature domain, the algorithm converges to samples belonging to the inner boundary, hence guaranteeing {\em optimality} conditions (the sample lies on the manifold DB). 
The driving and pooling system, informed by class data, enables the search path to point to regions of elevated sample density.
For all experiments, a pool of samples for each class has been used to sample the class reference in order to drive the algorithm's next step. 

\subsection{Proofs}
The algorithm convergence is guaranteed by the following proofs:
\subsubsection{Convergence direction}
Let NN $F^a: R^{n} \longrightarrow [0,1]$ be a continuous function, $x_0 \in C_0$ be a class 0 sample, and $x_1 \in C_1$ class 1 sample. The Intermediate value theorem asserts that:
\begin{equation*}
    F^a(x_0) = 0 < 0.5 < 1 = F^a(x_1)  \implies \exists x' \text{s.t.}\ F^a(x') = 0.5
\end{equation*}
Hence, we can define $x'$ from each pair $(x_0, x_1) \in C_0 \times C_1$:
\begin{equation*}
    \forall (x_0, x_1) \; \exists \lambda \in (0, 1) \; \text{s.t.} \; x' = \lambda x_0 + (1 - \lambda) x_1
    \text{ and } F^a(x') = 0.5.
\end{equation*}
So, by moving from one class to the other, we are crossing the DB at least once.
Sampling training data as targets ensures the possibility of reaching a trained DB portion.

\subsubsection{Algorithm halting}
By design, the magnitude of the next step is driven exclusively by the prediction distance to the target ($\approx 0.5$); thereby, the algorithm halts if and only if it reaches the DB.

By contradiction, suppose that the $magnitude = 0$ with $F^a(x) \neq 0.5$: 
\begin{align*}
magnitude = \| F^a(x) - 0.5 \|\\
0 = \| F^a(x) - 0.5 \|\implies F^a(x) = 0.5  
\end{align*}
Since assuming $F^a(x) \neq 0$ leads to a contradiction, our assumption must be false. Hence, the only possible solution is $F^a(x) = 0.5$.

\subsubsection{Looping}
Using low-size pools may lead to loops and so diverge. To cope with that, the algorithm implements a scaling function, but the user can also define it. The naive version implements a decreasing value.
\begin{align*}
    magnitude = magnitude * sf(step)\\
    sf(x) = \gamma^{x} \quad \text{with} \quad \gamma \in (0,1]
\end{align*}
The value of $\gamma$ tunes the magnitude according to the iteration of the algorithm to encourage big steps in the first phase of the search. With $\gamma = 1$, there is no effect.

\subsection{Validation of the proposed algorithm}
The proposed algorithm has been validated by considering four different scenarios in terms of simulated data distributions, and compared with two well-established algorithms allowing the DB detection and visualization through a bi-dimensional projection. 

The first two datasets were defined by two features, having different distributions in the feature space, that is: two classes and two clusters for each group (Custom) and a spiral shape (Spiral), respectively. Then, we increased the data complexity, relying on samples defined by three features for the third dataset and multiple features for the fourth, respectively.

Regarding the algorithms used for the comparison, we relied on SplineCAM (\cite{humayun2023splinecam}) and DeepView (\cite{schulz2019deepview}). The first was only employed in the two feature cases due to the inherent limitations of the algorithm in working with higher dimensional spaces. Conversely, DeepView allows processing feature spaces with higher dimensionality relying on UMAP \cite{mcinnes2018umap} for dimensionality reduction. In order to check the consistency of the outcomes of the three algorithms (IBS, SplineCam, and DeepView) in locating the DB, experiments were conducted on the two and three features cases, and the overlap of the respective boundary maps was assessed as described in the next Sessions.

\section{Materials and Methods}
\label{sec:mat_met}

\subsection{Datasets}
Synthetic datasets with two classes were generated for the experiments by composing Gaussian clusters, each one located around the vertices of a hypercube in a subspace of dimension $N$, where $I$ of the $N$ were the informative features. For each cluster, informative features were drawn independently from $\mathcal{N}(0, 1)$, and then randomly linearly combined within each cluster in order to add covariance. Finally, the clusters were placed on the vertices of the hypercube.

\paragraph*{Case 1 - two features}
Two datasets with two features were created. The first was a custom dataset with $2000$ samples. Both features were important for discriminating the two classes. Two clusters for each class were created. 

The second, called spiral dataset, defined in \cite{humayun2023splinecam}, was generated in order to assess the ability of our method to map the DB in a non-linear case. The number of samples used for creating this dataset was the same used in \cite{humayun2023splinecam}.

\paragraph*{Case 2 - three features}
A dataset with three features was generated with the same procedure described above (two feature cases). All the features of this dataset were relevant to the task.

\paragraph*{Case 3 - simulated brain view}
In order to generalize the proposed approach to an image context (multi-feature), a dataset that contains 2500 images belonging to two classes was generated. The number of features was set to $5290$, out of which $53$ were informative. 
After extracting the indices of the informative features, the data were fitted to a $109\times91$ brain mask, and finally, the images were processed with a Gaussian filter to locally spread the main features simulating local feature fingerprints in brain tissues.
\begin{figure}[!t]
    \centering
    \includegraphics[width=.5\linewidth]{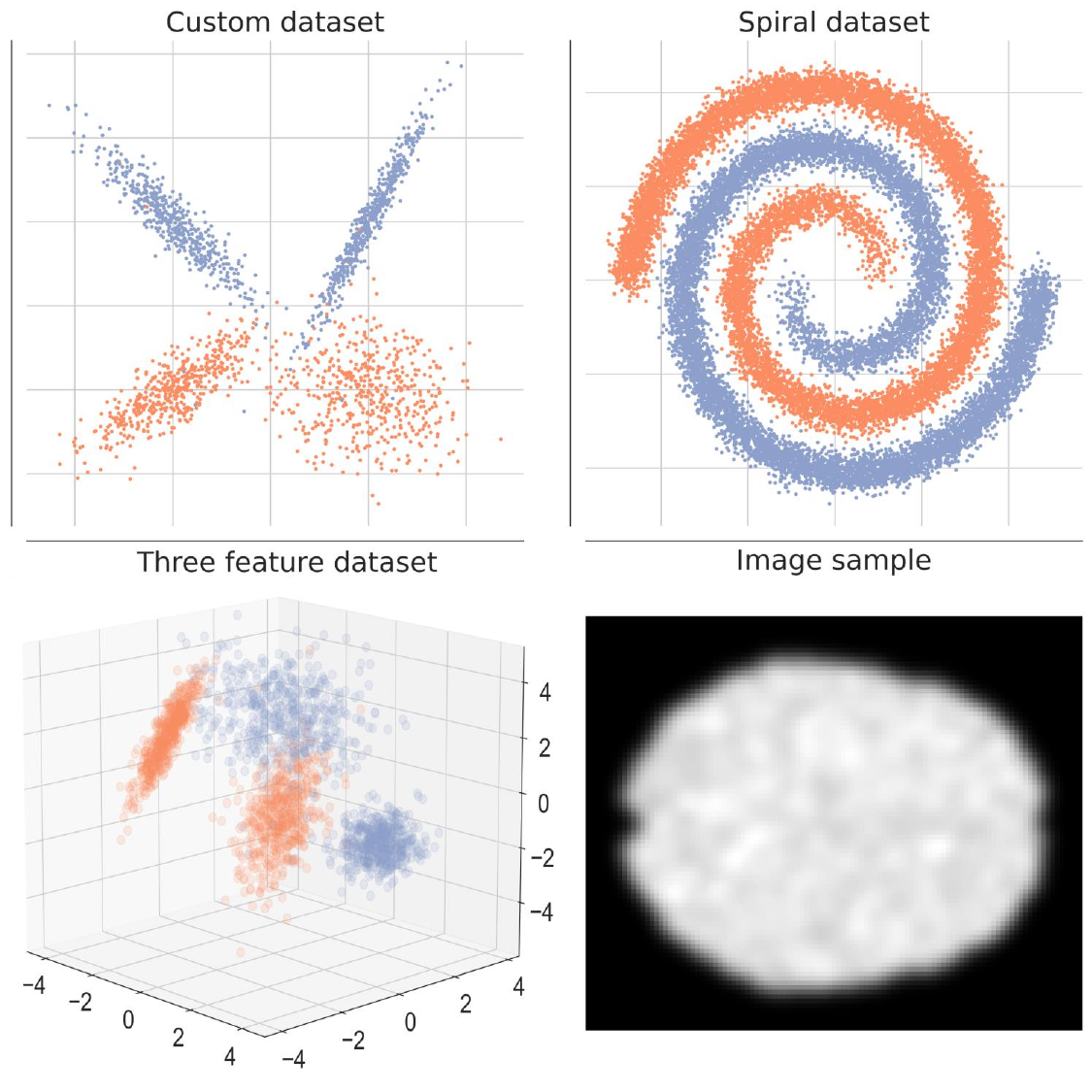}
    \caption{Dataset representation, starting from upper left \emph{Custom} dataset, then \emph{Spiral} dataset, and in the second row, \emph{Three} features dataset and finally an image from the \emph{Simulated brain views}. }
    \label{fig:datasets}
\end{figure}
Fig. \ref{fig:datasets} shows the generated datasets.

\subsection{Deep Learning Architectures}
We used Multilayer Perceptron (MLP) composed of five consecutive fully-connected layers with 10 neurons each and ReLU as activation function. The architecture was employed in Humayun et al. \cite{humayun2023splinecam} for the sake of the methods comparisons.
The output layer was defined by a single sigmoid neuron.
The network used for the image task was a CNN, composed of four convolutional layers (kernel size = $5\times5$ for the first two, and $3\times3$ for the last two), followed by a max-pooling layer after each pair, and then three fully-connected layers (number of nodes: $13248$, $256$, and $64$) followed by dropout layers. All the convolutional and fully-connected layers were activated by a ReLU function, except for the output one.

\paragraph*{Training and evaluation scheme}
Each dataset was initially split into $85-15$ percent of the samples for the training and testing sets, respectively. Then, each network was trained over the training set with a 10-fold cross-validation strategy for 15 epochs in order to avoid overfitting and a batch size of 128, and then tested on an external holdout test set.
Binary cross entropy was used as the loss function, and the Adam optimizer with a learning rate of $1 \times 10^{-3}$ was used for all cases, except for the synthetic image task, where a learning rate of $1 \times 10^{-4}$ and weight decay of $5 \times 10^{-6}$ were used. The performances are reported in Tab. \ref{tab:performances}.

\begin{table}[]
   \centering
   \caption{Classification performances on synthetic datasets.}
   \begin{tabular}{c|c|c}

        Dataset& Accuracy & F1 \\
        \textit{Custom} & $0.979 \pm 0.00014 $& $0.980 \pm 0.00001$ \\
        \textit{Spiral} & $0.995 \pm 0.00006$ & $0.995 \pm 0.00005 $\\
        \textit{Three features} & $0.979 \pm 0.00014$ &$ 0.976 \pm 0.00009$ \\
        \textit{Simulated brain} &$ 0.979 \pm 0.00046$ & $0.970 \pm 0.00050$ \\
   \end{tabular}
   
   \label{tab:performances}
\end{table}
\section{Results}
\label{sec:result}

This section presents the experimental results of this work, highlighting initially how different attributions can be extracted by IG using different BLs. Then, the comparison and validation of the proposed algorithm (IBS) with respect to SplineCAM and DeepView in sampling the DB is presented. Finally, an extended analysis of IG attributions related to the optimal baseline sampled from the DB is shown. 

\subsection{Baseline localization problem}
The baseline-related problem in computing the attribution in the IG method has been presented in Section \ref{sec:prob_def} where the optimal BL has been defined as a BL located inside the data distribution and on the DB. 

\begin{figure}[!t]
    \centering
    \includegraphics[width=\linewidth]{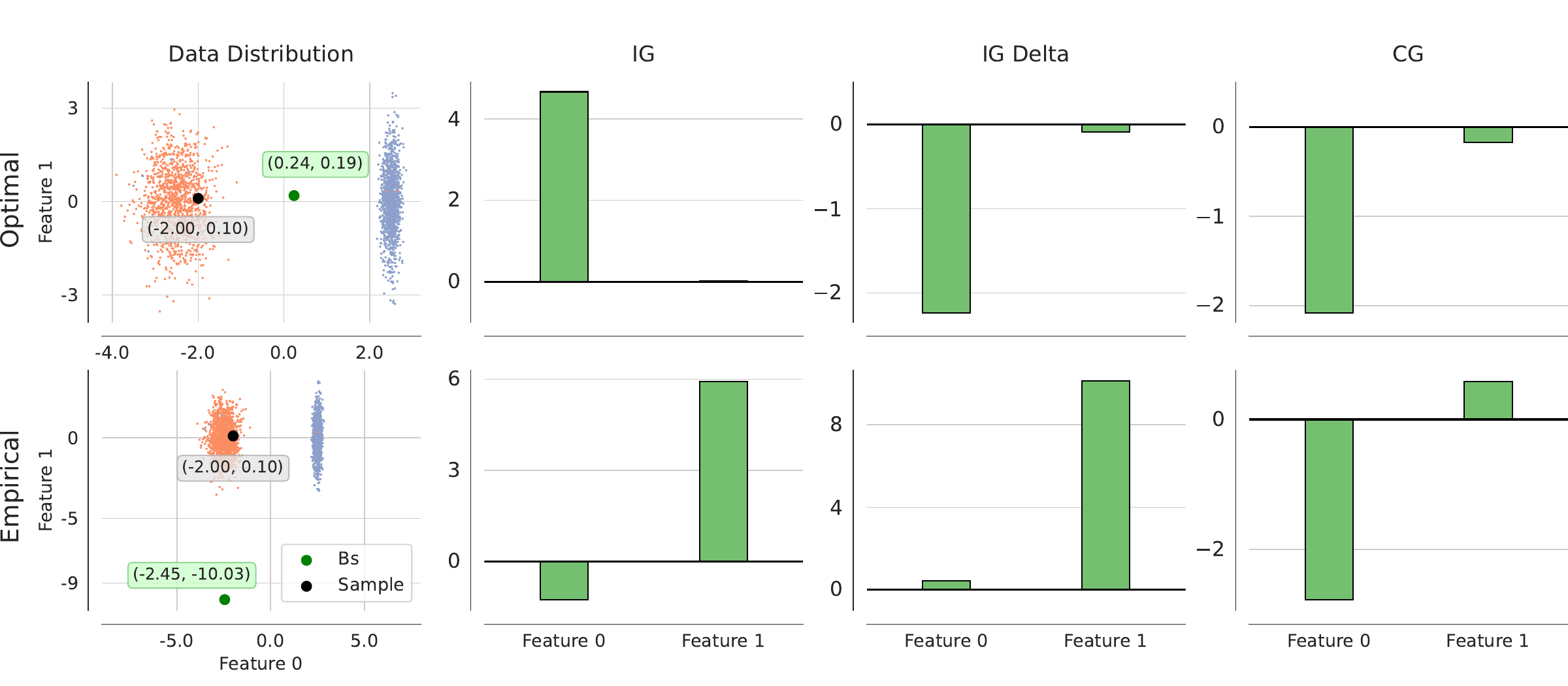}
    \caption{BL problem, with different BLs (First column, Feature 0 on X-axis and Feature 1 on Y-axis), the features attributions for the class 0 (Orange) are opposite (second column), then \emph{Delta} (third column), and finally in the last column are reported the \emph{Cumulative Gradients} (CG).}
    \label{fig:in_out_2}
\end{figure}

\begin{figure}[!t]
    \centering
    \includegraphics[width=\linewidth]{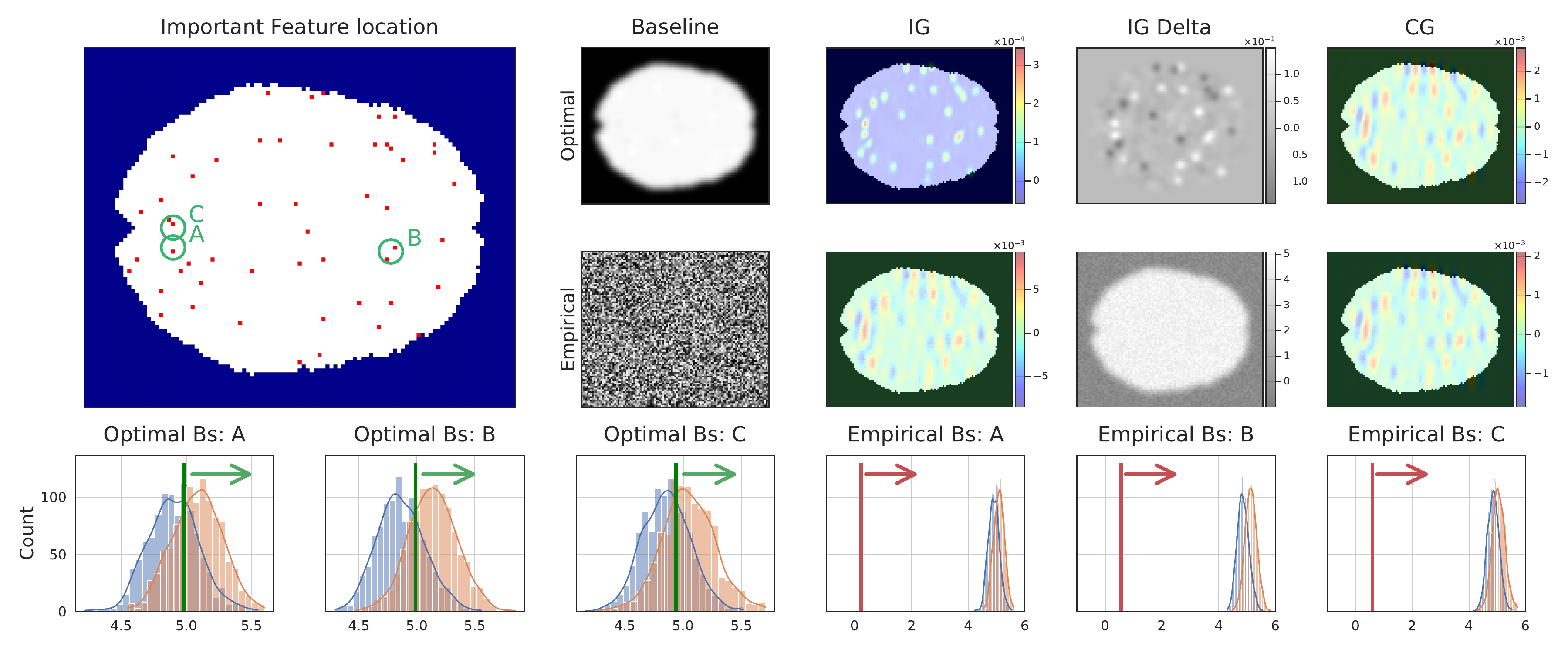}
    \caption{Image case. Left image: important feature locations labeled as A, B, and C. First and second row: comparison between two BLs on IG attributions (similar to the previous figure). Bottom row: data distributions of the three highlighted features (A, B, C), BL location, and gradient directions. The green color is used for optimal BL and red for the random one. }
    \label{fig:brain_attr_cd}
\end{figure}

Fig. \ref{fig:in_out_2} and Fig. \ref{fig:brain_attr_cd} show experimental results on two synthetic datasets, two features case (Case 1) and simulated images (Case 3), respectively, showing how an internal and external BL result in different attributions. In particular, Fig. \ref{fig:in_out_2} shows that IG with the internal BL (top row) results in positive attribution values, because of the product between the CG and Delta. 
On the other hand, in the case where the baseline was outside the data distribution (second raw), the IG attribution had both positive and negative values. 

Figure \ref{fig:brain_attr_cd} illustrates Case 3, that is the synthetic image case study. The upper-left image shows the location of the most important features (pixels in this case) where three, labelled as A, B, and C, were highlighted for the subsequent analysis (bottom row). The first and second upper-right rows show the BL, IG attributions, the Delta, and the CG for the optimal and random BLs, respectively. The IG attributions extracted from the optimal BL had all positive values as expected, in contrast to the random one for which both positive and negative values can be observed. The Delta of the optimal BL showed a more faithful representation of the problem under analysis.
The bottom row of Fig. \ref{fig:brain_attr_cd} represents the features' distributions of the pixels labeled as A, B, and C, as well as the gradient direction for the optimal (first three plots) and arbitrary (last three) BLs, respectively, for the orange class. For the optimal BL (green bar and arrow, respectively) the gradient points in the correct direction.


\subsection{Decision boundary search and IBS algorithm validation}
In what follows, the ability of the proposed algorithm to find the DB in all the datasets is shown, also validating the outcome with SplineCAM and DeepView algorithms.

\begin{figure}
    \centering
    \begin{subfigure}[!t]{.48\linewidth}
        \centering
        \includegraphics[width=\linewidth]{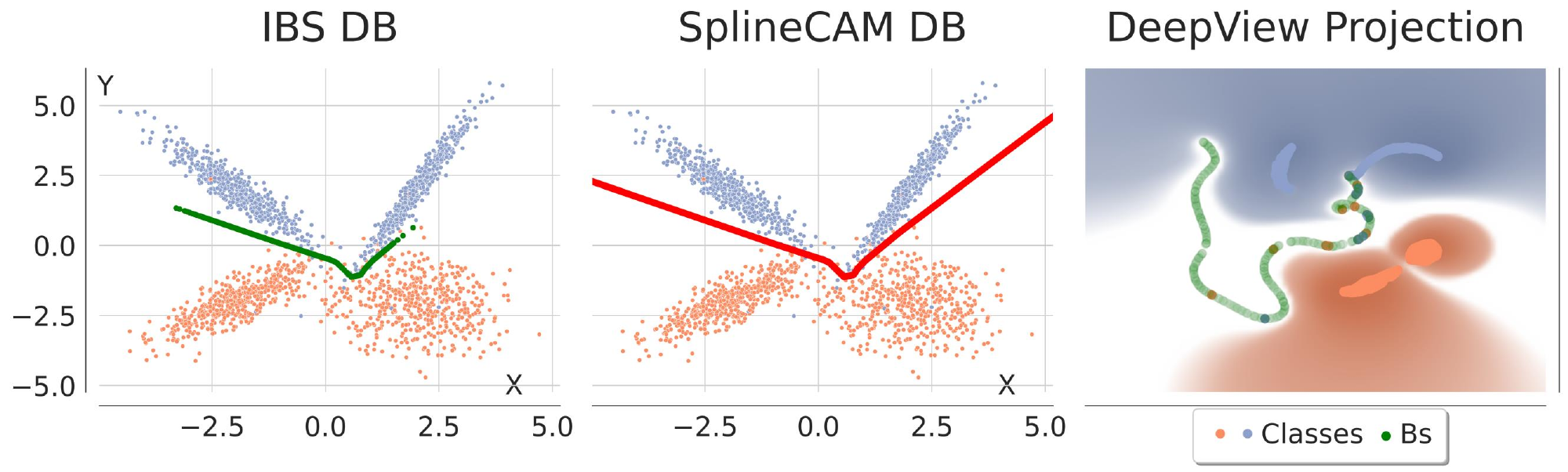}
        \caption{\emph{Custom} dataset. The first and second subfigures represent the data and the detected boundary, in green for IBS and in red for SplineCAM. The lust one represents the DeepView outcome, where the orange and the blue represent the class influence area, and the white indicates the boundary area.}
        \label{fig:custom_r}
    \end{subfigure}
    \quad
    \begin{subfigure}[!t]{.48\linewidth}
        \centering
        \includegraphics[width=\linewidth]{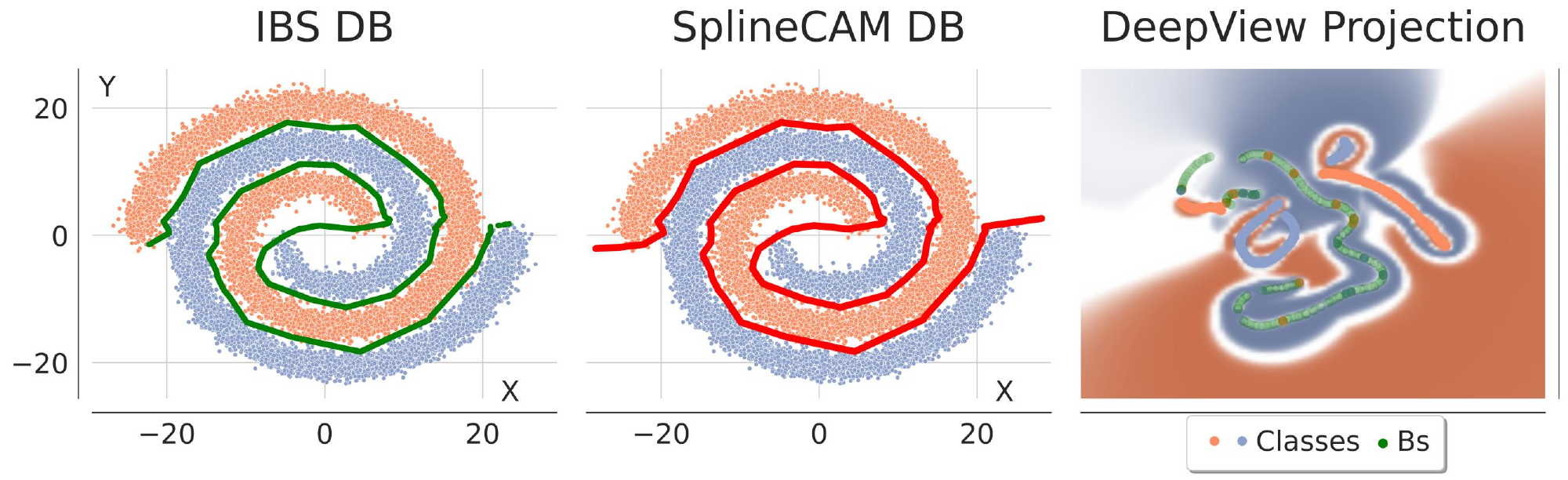}
        \caption{\emph{Spiral} dataset. Data distribution and boundary resulting from IBS (green), SplineCAM (red), and DeepView projection, where the orange and the blue indicate the class influence and white space the boundary areas.}
        \label{fig:spiral_r}
    \end{subfigure}
        
\end{figure}

Fig. \ref{fig:custom_r} shows the data distribution of the custom dataset overlayed to the DB as detected by the different methods. The first image (left) highlights in green the DB identified by the proposed algorithm, while the second (middle) and third (right) images show the SplineCAM BD in red, and DeepView projection where the orange and white regions represent the class influence and boundary areas, respectively.
As it can be observed, the outcomes of IBS and SplineCAM overlap exactly and both lie in the inner space. In the DeepView projection space the IBS DB is correctly placed in the white area. This provides evidence of the consistency of the results across methods. 

Fig. \ref{fig:spiral_r} illustrates the case of the spiral dataset. Again, IBS detects the same DB as SplineCAM. This example highlights a limitation of the DeepView algorithm, that was not able to correctly locate the boundary. 

\begin{figure}
    \centering
    \begin{subfigure}[!t]{.48\linewidth}
        \centering
    \includegraphics[width=1\linewidth]{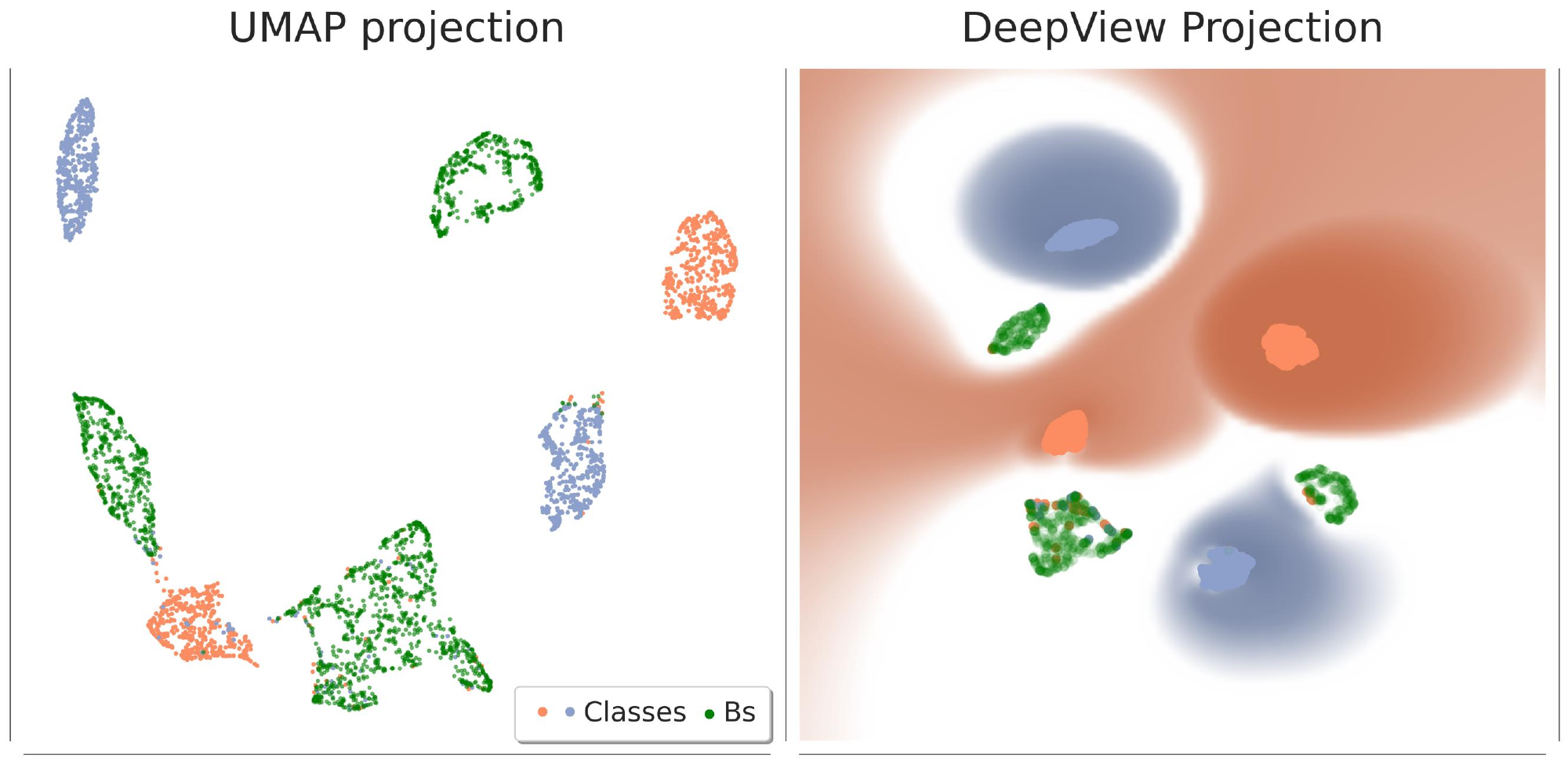}
    \caption{\emph{Three feature} dataset. The left figure represents the data distributions and the IBS BLs extracted as a UMAP projection for the sake of visualization (UMAP preserves the clusters), the right one regards DeepView projection, and the colored area represents the class influence area; meanwhile, the white space indicates the boundary. }
    \label{fig:3f_r}
    \end{subfigure}
    \quad
    \begin{subfigure}[!t]{.48\linewidth}
        \centering
    \includegraphics[width=\linewidth]{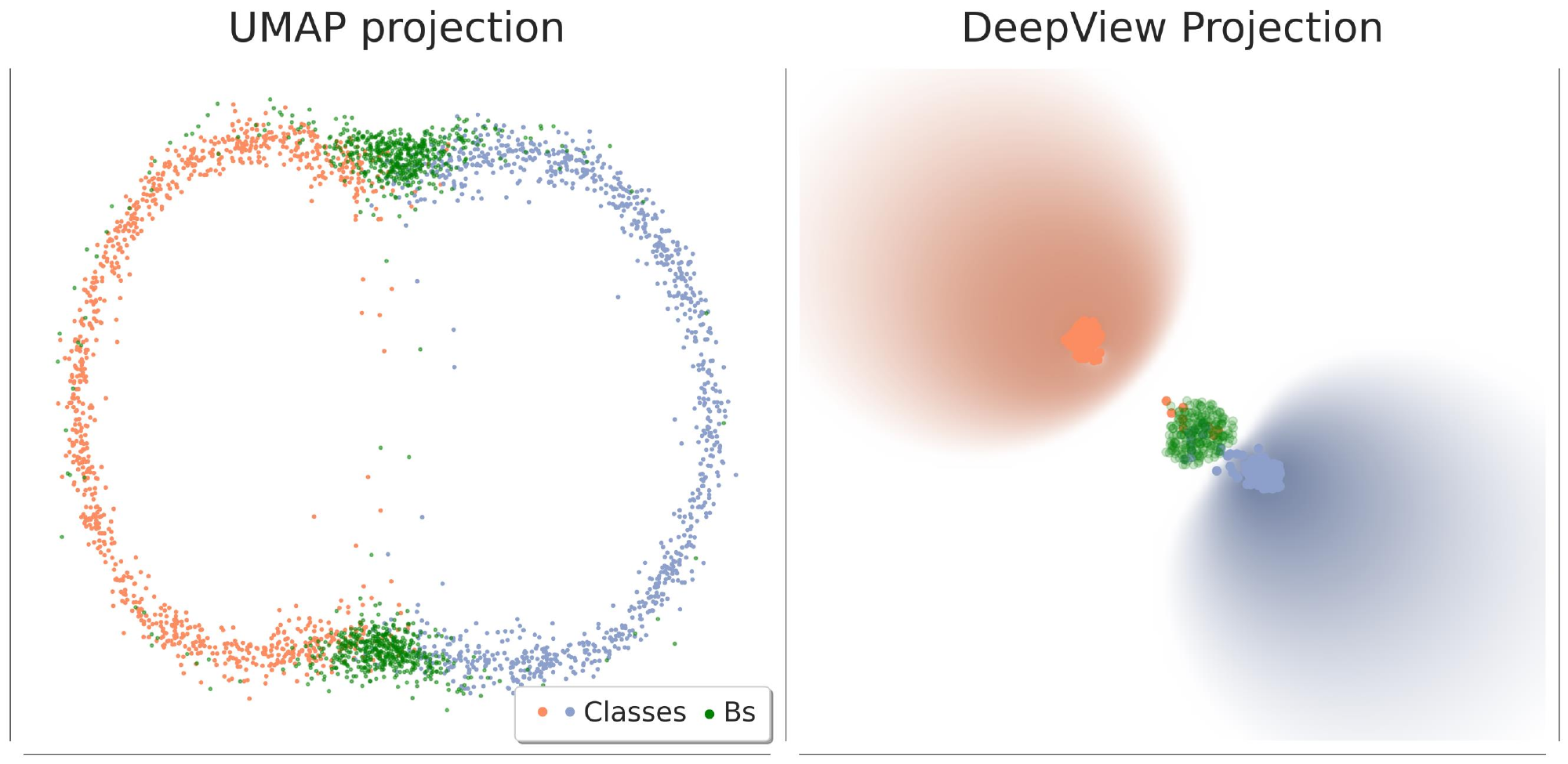}
    \caption{\emph{Simulated brain view} dataset. The left figure represents the data distributions and the IBS BLs extracted as a UMAP projection for the sake of visualization, while the right one regards DeepView projection where the colored area represents the class influence area and the white space indicates the boundary.}
    \label{fig:multi_res}
    \end{subfigure}
        
\end{figure}

Fig. \ref{fig:3f_r} shows the results for the three features dataset (Case 2). For the sake of visualization, the data distributions with the IBS BLs are shown as a two-dimensional projection performed by UMAP.
Of note, UMAP preserves the clusters of the original space.
The left image of Fig. \ref{fig:3f_r} shows the UMAP projection of the data distribution with the BLs found by the proposed algorithm, while the right image represents the DeepView projection that confirms the location of the BLs in the DB.
On this dataset, the SplineCAM validation was not performed due to its limitations to two features.


Fig. \ref{fig:multi_res} shows the results for the synthetic brain dataset (Case 3). 
The left image shows the projection in a two features space of the image classes and the set of BLs found by the proposed IBS algorithm, while the right image shows the DeepView projection of the images and the BLs, highlighting how the BLs reside in the white background (the DB) thus confirming the consistency of the DB position across methods.

\subsection{Optimal baseline standard validation}
\begin{figure*}[!t]
    \centering
    \includegraphics[width=\linewidth]{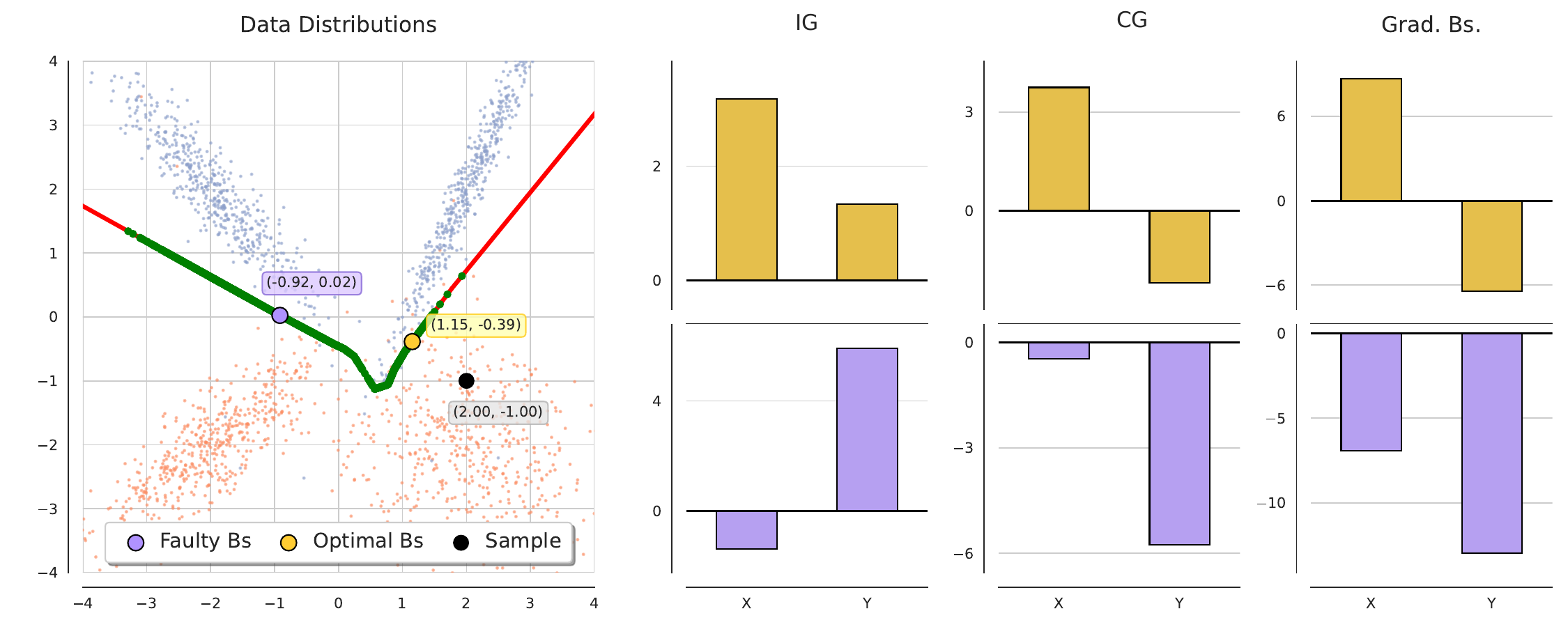}
    \caption{\emph{Custom}: different BL attributions and gradients, the first represents the class distributions, boundary (red and green), bs chosen (purple and yellow), and sample (black dot). The first column represents the IG attributions for the two BLs, the second regards the \emph{Cumulative Gradients}, and the last one represents the network gradient with BL as input.}
    \label{fig:bs_Attr_gr}
\end{figure*}
\begin{figure*}[!t]
    \centering
    \includegraphics[width=1\linewidth]{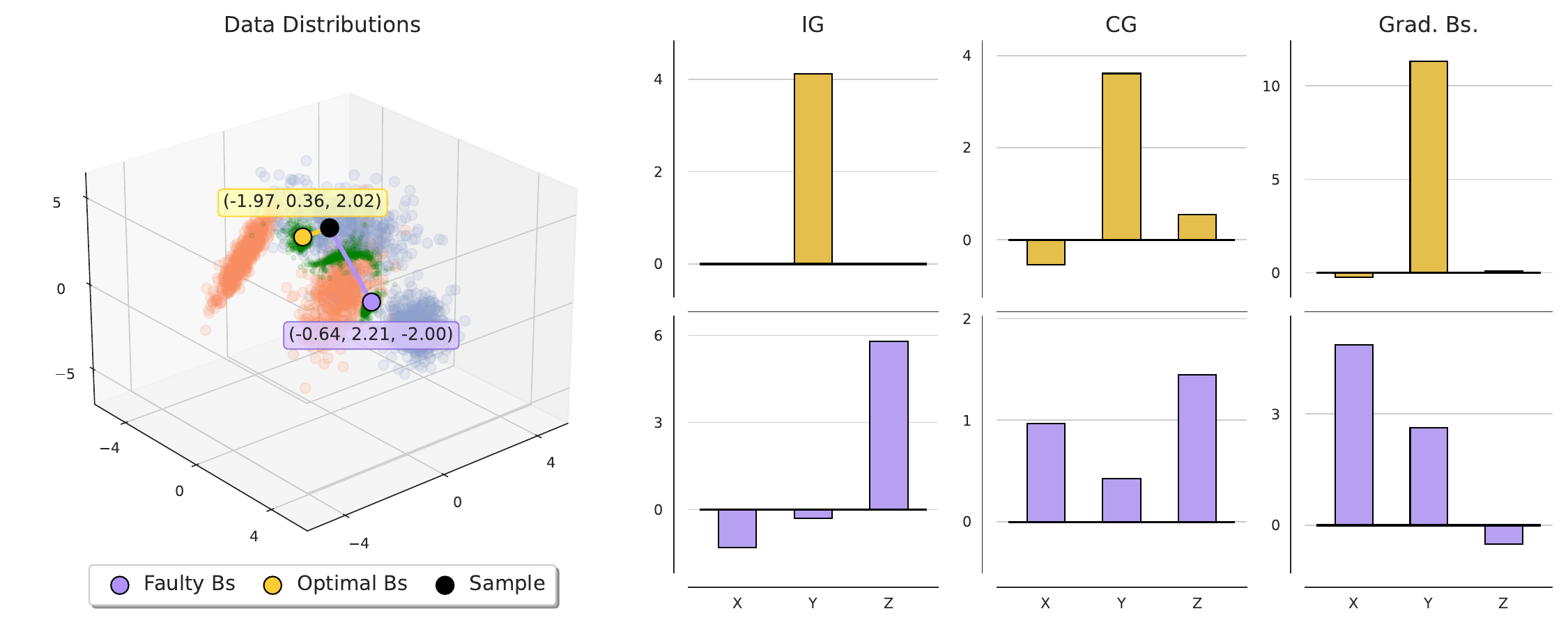}
    \caption{\emph{Three feature}, different BL attributions, \emph{Cumulative Gradients} and BL gradients. The first plot from the left concerns the data distribution with IBS boundary (green clusters) and chosen BLs (yellow, purple). The first column concerns the \emph{Integrated Gradients}, the second column represents \emph{Cumulative Gradients}, and the last one contains the BL gradient attribution.}
    \label{fig:ig_vs_sal_3}
\end{figure*}
Fig. \ref{fig:bs_Attr_gr} shows the different attributions obtained using two different BLs extracted from the proposed algorithm, respectively representing the optimal BL (orthogonal projection of sample) and a faulty BL (far from the sample). The left image shows the class distributions, highlighting the DB BLs extracted from the proposed algorithm and the SplineCAM boundary, in addition to the two selected BLs (yellow and purple dots for the optimal and faulty BLs, respectively) and the investigated sample (black dot).
On the right side, the resulting attributions from the two different BLs are displayed, showing IG, CG, and the gradients at the BL (saliency), in the first, second, and third columns, respectively. The color links the BL with its attributions.

Differences can be observed in the gradients computed relying on the two BLs (third column). In the case of the purple BL, the gradient computation has been corrupted by the multiple crossings of the path to the sample with the DB, as discussed in Fig. \ref{fig:grad_comp}.

In consequence, the IG attributions relying on the optimal BL and reported in the first column take only positive values, while both positive and negative scores are obtained for the faulty one.

The same analysis is proposed for the three features case (Case 2). Fig. \ref{fig:ig_vs_sal_3} shows on the left the data distributions. Like in Fig. \ref{fig:bs_Attr_gr}, the optimal and faulty BLs are the yellow and purple dots, respectively, while the black dot is the sample under analysis. A similar trend is observed in the attributions, with the only difference of three features. Only positive scores are extracted from IG with the optimal BL, and a concordance in the sign of the CG gradients and the gradients at the BLs is reported, in contrast to the attributions extracted with a faulty BL where the gradients are influenced by different DB segments, hence corrupting the attributions. 

\section{Discussion}
\label{sec:discussion}
In this Section, the IBS algorithm is discussed as well as its validation with respect to the SOA. 
Then, the main limitations and open issues of the proposed algorithm are presented and the main contributions are summarized. 

\subsection{IBS identification and validation}
\label{SS1}
The IBS algorithm has been developed with the objective of extracting domain- and task-specific BLs in order to compute reliable and accurate feature attributions. The algorithm extracts the optimal BL from the DB of the network, hence a neutral signal, that is located in the inner portion of the boundary thanks to the guided-search procedure.
Our experimental results demonstrated that the BL computed with the proposed algorithm extracted only positive attribution, hence balancing the negative value in the CG with the most appropriate IG Delta factor.
Some works suggested solving this problem using only the absolute value of the attributions \cite{hooker2019benchmark, sturmfels2020visualizing}, but limiting the interpretability of the outcomes and the possibility to link the sign of the attribution to the modulations of the features driving the sample to a specific class.

Previous works have already considered the idea of using samples of the DB as a baseline, but with limited interpretations and applications. Izzo et al. \cite{izzo2020baseline} proposed the use of the DB samples as BLs, but with a different intuition. They assumed that a neutral value can lead to a point in the input domain that could be used as a BL, proposing an inverse function of the Single Layer Perceptron (SLP), capable of reconstructing the input given the output of the layer. The main limitation of their approach was that it was limited only to the SLP, and they did not propose an affordable algorithm to deal with multilayer architectures. 

Additionally, our experimental results and analysis moved the focus to the concept of an optimal BL, which carried some properties, and using the provided guidelines, it was possible to set up a post-hoc XAI analysis that extracted reliable and accurate attributions and knowledge from the NN.
In detail, the results demonstrate that the optimal BL, defined as the closest point on the DB, was suitable for obtaining the most accurate feature relevance in different scenarios using the IG method. In contrast, when a BL far from the sample was selected, the attributions were corrupted due to the relation between gradient and orientation of the DB segments, explained in Lee et al. \cite{lee1997decision}. This situation implies the non-uniqueness of attributions in the presence of a non-linear boundary. Rather, it creates different attributions based on the number of segments with different orientations, hence affecting the CG since it computes the gradients several times along the path from the BL to the sample, with the possibility of crossing other DB segments, and thus further corrupting the final attributions.
Furthermore, an optimal BL brings with it the ability to identify the location of a specific class by exploiting its own gradient and its features. With that, it is also possible to provide a counterfactual explanation of the sample.

The experimental results showed that the proposed IBS algorithm sampled the DB in both low- and high-dimensional cases, also including complex scenarios.
In the low feature space (two features, Case 1), the results were close to or equal to those of SplineCAM. However, SplineCAM suffers from many constraints that limit its applicability to specific models and hyperparameter choices. 
As previously discussed, there is no interest in that space (outside the data distribution) since it was not well-trained \cite{karimi2019characterizing, lee1997decision, arras2022clevr}. Conversely, IBS was directed to the inner DB in the data domain, hence enabling the proposed algorithm to extract a reliable DB from which the optimal BL can be sampled.
In the high-dimensional spaces (three features and images), SplineCAM validation was not performed due to its limitations since it allows investigating only two dimensions and also because SplineCAM was not able to manage all networks' layers, such as max-pooling.
In this case, DeepView was employed to validate the BL extracted from IBS by projecting the data in a bidimensional space, allowing to observe the difference across classes and the DB in both low (two features)-and high (three features and images)-dimensional spaces.
Considering the experimental results obtained with this algorithm, the computed baselines were correctly placed on the white regions of DeepView, indicating the DB. Similarly to SplineCAM, DeepView performed the feature reduction using UMAP, hence preserving the cluster localization. Considering this feature of DeepView, it can be observed that in the obtained results, the BLs are localized in-between groups, hence representing the inner DB.

Note that what emerged from the experimental results highlighted the DB spatial orientation and corresponding class separation. The orientation of the division tied in perfectly with what was shown in \cite{lee1993feature}, where the authors linked feature attributions to the orientation of the DB.

\subsection{Main contributions and findings}
\label{SS2}
The experimental results showed the importance of selecting an optimal BL, identified as the orthogonal projection of the investigated sample on the inner DB, for computing the attributions through an XAI BAM method. To this end, the IBS algorithm has been proposed. Allowing to detect the inner DB in the feature domain, using the samples with similar distributions that constitute the dataset under analysis.
If this optimal BL cannot be detected, it is suggested to use the closest one, even if the final attributions will correspond to a sub-optimal solution.
Additionally, the ability of the proposed algorithm to sample the DB has been validated using two SOA methods, providing evidence of the consistency of the outcomes across methods while highlighting the inherent limitations of SplineCam and DeepView, such as a limited number of manageable layers for the first, and computing the projection to reduce the dimensionality for the second, respectively.
The IBS is also proposed in a GPU-friendly version, which allows faster computation and, hence, a lower carbon footprint. 
Finally, this work raises a warning in the choice of the BL; a BL far from the sample under analysis, even if residing in the DB, could generate corrupted attribution due to the influence of the different DB segments.

\subsection{Limitations and future directions}
\label{SS4}
The main limitation of this work lies in the scope: only the IG method was considered, and performance evaluation relied solely on synthetic data. However, the framework is open and flexible and can be naturally generalized to other baseline-guided methods and realistic scenarios. 
Accordingly, future directions include expanding the analysis to other BAM methods relying on biomedical datasets. In addition, the proposed optimal BL detection method can be exploited for probing robustness and performing adversarial sampling, opening the way to knowledge distillation and model quantization.

\section{Conclusion}
\label{sec:conc}
In this work, we performed an extended analysis and defined some guidelines to represent the optimal BL that has to be used in the BAM XAI methods, with a particular focus on the well-known Integrated Gradients method. For computing the optimal BL, we proposed the IBS algorithm that is able to sample the inner DB of a NN in order to extract the BL. Experimental results and validation procedures demonstrated the ability of our algorithm in extracting BLs that computed reliable and accurate attributions in conjunction with IG with respect to empirical BLs.

\bibliographystyle{habbrv.bst}
\bibliography{main}

\end{document}